\documentclass{sig-alternate-05-2015}
\usepackage{tabularx}
\usepackage[draft]{hyperref}
\usepackage{graphicx}
\usepackage{pgfplots}
\usepackage{algorithmicx}
\usepackage{algorithm}
\usepackage{algpseudocode}
\usepackage{pgfplotstable}
\usepackage{booktabs} 
\usepackage{longtable}
\usepackage{float}
\usepackage{url}
\usepackage{tabularx}
\usepackage[utf8]{inputenc}
\usepackage[caption = false]{subfig}
\newcounter{magicrownumbers}
\newcommand\rownumber{\stepcounter{magicrownumbers}\arabic{magicrownumbers}}

\makeatletter
\def\@copyrightspace{\relax}
\makeatother

\begin{document}

\title{Social media mining for identification and exploration of health-related information from pregnant women 
}

\numberofauthors{4} 

\author{
\alignauthor
Pramod Chandrashekar\\
      \affaddr{Department of Biomedical Informatics}\\
      \affaddr{Arizona State University}\\
      \affaddr{Tempe, AZ, USA}\\
      \email{\url{pbchandr@asu.edu}}
\alignauthor
Arjun Magge\\
      \affaddr{Department of Biomedical Informatics}\\
      \affaddr{Arizona State University}\\
      \affaddr{Tempe, AZ, USA}\\
      \email{\url{amaggera@asu.edu}}
\alignauthor Abeed Sarker\\
      \affaddr{Department of Biostatistics and Epidemiology}\\
      \affaddr{University of Pennsylvania}\\
      \affaddr{Philadelphia, PA, USA}\\
      \email{\url{abeed@upenn.edu}}
\and  
\alignauthor Graciela Gonzalez\\
      \affaddr{Department of Biostatistics and Epidemiology}\\
      \affaddr{University of Pennsylvania}\\
      \affaddr{Philadelphia, PA, USA}\\
      \email{\url{gragon@upenn.edu}}
}

\maketitle
\begin{abstract}

Widespread use of social media has led to the generation of substantial amounts of information about individuals, including health-related information. Thus, social media provides the opportunity to study health-related information about selected population groups who may be of interest for a particular study. In this paper, we explore the possibility of utilizing social media data to perform targeted data collection and analysis from a particular population group--- pregnant women. We hypothesize that we can use social media to identify cohorts of pregnant women and follow them over time to analyze crucial health-related information. To identify potentially pregnant women, we employ simple rule-based searches that attempt to detect pregnancy announcements with moderate precision. To further filter out false positives and noise, we employ a supervised classifier using a small number of hand-annotated data. Following the identification of a reasonably sized cohort, we collect their posts over time to create longitudinal health timelines and attempt to divide the timelines into different pregnancy trimesters. Finally, we assess the usefulness of the timelines by performing a preliminary analysis to estimate drug intake patterns of our cohort at different trimesters. Our rule-based cohort identification technique collected 53,820 users over thirty months from Twitter. Our pregnancy announcement classification technique achieved an F-measure of 0.81 for the pregnancy class, resulting in 34,895 user timelines. Analysis of the timelines revealed that pertinent health-related information, such as drug-intake and adverse reactions can be mined from the data. Our approach to using user timelines in this fashion has produced very encouraging results and can be employed for an array of other important tasks where cohorts, for which health-related information may not be available from other sources, are required to be followed over time to derive population-based estimates.

\end{abstract}

\section{Introduction}

Pre-market clinical trials assess the safety of drugs/medications (we use the terms interchangeably in this paper) in limited settings, and so the effects of those drugs on particular cohorts (\emph{e.g.}, pregnant women, children, or people suffering from specific conditions) cannot be assessed. Spontaneous reporting systems, such as the FDA Adverse Event Reporting System (FAERS), are used for post-marketing drug safety surveillance and they provide a mechanism for reporting adverse events associated with medication consumption. Although these sources may accumulate drug safety knowledge about specific population groups, studies have shown that they suffer from various problems, such as under-reporting \cite{harpaz2012novel}. To overcome these problems, additional sources of information are being actively utilized for pharmacovigilance tasks. Studies have shown that 26\% of online adults discuss health information using social media \cite{businesswire2012twenty}, with approximately 90\% women using online media for health-care information, and 60\% using pregnancy related apps for support. These statistics suggest that social media sources may contain key information regarding specific cohorts, such as pregnant women, and their drug usage habits.




Although consuming drugs during pregnancy is not recommended by doctors worldwide, their usage is commonplace. For example, during pregnancy, women continue taking prescription drugs for ailments which preceded the pregnancy. For common health problems like heartburn, common cold and body pains, women tend to take over-the-counter medicines which may cause harm to the fetus. Past research has also indicated that 50\% of the pregnancies in the United States are unintended \cite{finer2006disparities}. In such cases, the fetus may be exposed to drugs without the mother's explicit knowledge. Currently, the U.S FDA maintains a list of pregnancy exposure registries\footnote{\url{http://www.fda.gov/ScienceResearch/SpecialTopics/WomensHealthResearch/ucm134848.htm}} that collect health information on exposure to medical products during pregnancy. Such registries require pregnant women to voluntarily sign up, and hence, they suffer from low enrollment and follow-up rates \cite{sinclair2014advantages}. Considering the fact that infant mortality rates are estimated to be at 5.96 deaths per 1,000 live births \cite{march2013cdc}, and that the causes of 50\% of these birth defects are unknown \cite{lobo2008birth}, identifying and utilizing additional sources for monitoring health information of pregnant women, such as social media, is of paramount importance.

A very popular social network, that is currently being extensively used for public health monitoring tasks, is Twitter--- a micro-blogging site which is actively used by over 313 million users.\footnote{\url{https://about.twitter.com/company}. Accessed on: 03/03/2016.} Despite the noisy nature of data on Twitter, because of the high volume and frequency, it is an attractive resource for big data mining tasks. In addition to widely used social networks like Twitter, there are also \emph{online health communities}, which facilitate health-related information sharing over the Internet. One such online health community is DailyStrength\footnote{\url{http://www.dailystrength.org}. Accessed on: 03/03/2016.}, which has over 400,000 members engaging in discussions among its 500+ groups. In contrast to tweets, the posts in online health forums like DailyStrength have no strict constraints on word counts. The language used is more formal and the availability of domain-specific discussion forums increase the chances of finding relevant medical information from discussions \cite{o2013survey}. Thus, Twitter and DailyStrength present quite different types of social media chatter, with the data from the latter being significantly lower in terms of both volume and noise. Both these data sources carry health-related knowledge expressed by various cohorts but require customized techniques for mining the knowledge encapsulated.

\subsection{Motivation, Goals and Contributions}

Given the limited amount of information that is available about pregnant women during pre-market clinical trials, there is a need to explore additional resources of information. The presence of large amounts of social media data, which hold crucial health-related information, presents a strong motivation for developing frameworks for mining longitudinal information from this resource. Based on these motivations, the goals of this paper are as follows:
\begin{itemize}

\item Develop natural language processing (NLP), machine learning, and information retrieval (IR) methods for accurately identifying a cohort of pregnant women and collecting their social media timelines.
\item Perform preliminary analyses of the extracted health timelines to assess their usefulness, identify limitations, and establish future research goals. 
\end{itemize}

The main contributions of the paper are as follows:
\begin{itemize}
\item We present a framework by which social media data can be used to identify and collect information about pregnant women. 
\item We show that health timelines collected from social media contain crucial health-related information, which may be used in longitudinal studies.
\item We discuss techniques for further dividing the timelines into pregnancy trimesters and verify that trimester-specific information can also be mined from the timelines.
\item We discuss the current limitations of our novel idea and outline future directions
\end{itemize}


The rest of the paper is organized as follows: in Section 2, we briefly outline past research related to ours, including social media mining and data-centric approaches to pregnancy-safety monitoring; in Section 3, we detail our methods for identifying cohorts from the two social media sources, extracting relevant longitudinal data, and analyzing the data in a preliminary fashion; in Section 4, we present our results and provide discussions regarding our plans to build on this pilot for larger future projects; in Section 5, we discuss the limitations of our work and outline some planned future work; and we conclude the paper in section 6. 

\section{Related Work}
Research work most closely related to ours is in the domain of pharmacovigilance from social media, although, to the best of our knowledge, no past research has attempted to identify and follow longitudinal cohort information from this domain. Most of the research in pharmacovigilance and drug safety surveillance has focused on identifying adverse reactions associated with medications. Some past research has attempted to employ classification techniques to determine adverse drug reaction (ADR) assertive posts. For these tasks, two primary techniques have been attempted: lexicon-based classification and supervised classification. In lexicon-based classifications \cite{nikfarjam2011pattern,leaman2010towards} a given text is classified as having an ADR if it meets a set of specified lexical rules. Supervised classification techniques \cite{sarker2015portable,bian2012towards} involve training classifiers using features from annotated data (used as training data) to automatically make classification decisions on test data based on observed probabilities in the training data. 


Due to the advances in natural language processing (NLP) and data science techniques, social media has recently been used for a variety of public health monitoring tasks in addition to pharmacovigilance \cite{paul2016psb}. These include monitoring the patterns of influenza \cite{aramaki2011twitter}, tracking tropical diseases like dengue fever \cite{gomide2011dengue}, and analyzing disease outbreaks such as E. coli \cite{diaz2012tracking} and Ebola \cite{odlum2015twitter}. In behavioral medicine research, social media has been used to study users' lifestyle and analyzing the health-related choices they make. Researchers have used social media to study nutrition \cite{sharma2015detecting} and obesity patterns \cite{mejova2015foodporn}. Applications also include analyzing alcohol \cite{aphinyanaphongs2014text}, nicotine \cite{prier2011identifying}, and drug abuse \cite{genes2014twitter}.There has been some research in timeline creation \cite{li2014timeline} and event extractions from timelines \cite{wen2013extracting,li2014major,choudhury2014personal} for specific events. However, little effort has been invested in generating the summary of health-related data.

Only a handful of studies has attempted to predict pregnancy outcomes using quantitative data. Ines Banjari \emph{et al.} \cite{banjari2015cluster} used clustering on a collection of questionnaire results accompanied by blood samples of 222 pregnant women who were in the first trimester. The authors performed hierarchical clustering considering three main features namely pre-pregnancy BMI, their age, and hemoglobin content. Via cluster analysis, the authors found that women with higher pre-pregnancy BMI and age have higher risks of complications during pregnancy. Laopaiboon \emph{et al.} \cite{laopaiboon2014advanced} studied the effect of maternal age and pregnancy outcome using health records of 308,149 singleton pregnant women. They used a multilevel, multivariate logistic regression with clustering technique to perform the study and found that 12.3\% of these women had advanced maternal age (AMA) which varied across countries. Von Mandach \emph{et al.} \cite{wettach2013pharmacovigilance} studied 202 fetal disorders from Swiss ADR database using records classified by regional pharmacovigilance centers as having ADRs. They performed a likelihood ratio and t-test and found that fetal disorders were closely associated with the ADRs of drugs they consumed. All these pregnancy-related studies have involved data sources from clinical records, reports, hospital patient data which often is expensive to obtain. Also, little information is available on lifestyle habits and drug usage after the patient's exit the medical facilities. Hence, social media and health forums are potentially attractive sources for extracting health information, drug usage patterns and their effects. Small samples of social media data have been used for performing pregnancy-related studies--- such as the work by De Choudhury \emph{et al.} \cite{de2013predicting}, where 376 women were monitored to predict postpartum changes. Automatically collecting and processing large samples of social media data, however, presents significant challenges due to the lack of structure and use of informal language \cite{sarker2015utilizing}.

 \section{Method}
\begin{figure*}
\centering
    \includegraphics[width=.99\textwidth]{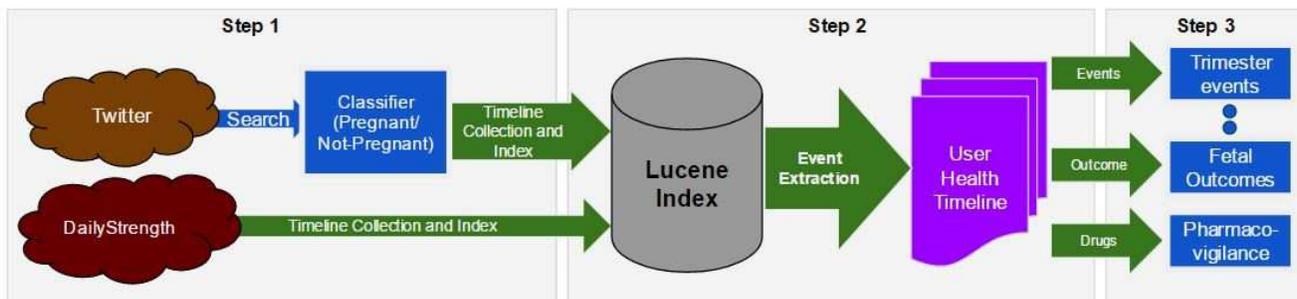}
    \caption{System architecture depicting our social media mining pipeline collecting and analyzing cohorts from social media.}
    \label{fig:sysArch1}
\end{figure*}

Figure \ref{fig:sysArch1} gives a detailed illustration of our proposed system, which is broadly divided into three main steps: Data Collection and Classification, User Health Timeline Extraction, and Timeline Analysis. For data collection, we discuss how cohort timelines can be collected from the differing interfaces of Twitter and DailyStrength. For the last step of the analysis, we show how the timelines can be divided into pregnancy trimesters so that trimester-specific information, such as drug usage, can be further analyzed. Each of these steps is detailed in the following subsections. 

\subsection{Data Collection and Classification}
Twitter and DailyStrength are the sources of our data for this study. We collected tweets originating from women announcing their pregnancy during a thirty-month time period, from January 2014 to September 2016. To identify our initial set of \emph{potentially pregnant} women, we applied simple search expressions of the forms \textit{``i'm * weeks/months pregnant"} and \textit{``i am * weeks/months pregnant"}, with minor variations adding to 18 queries. DailyStrength has a very different structure compared to Twitter, and the website is divided into individual forums for specific cohorts. We obtained our data from five forums on DailyStrength (Pregnancy, Pregnancy After Loss Or Infertility, Pregnancy Teens, Stillbirth, and Miscarriage). We collected all the posts from all the users from these forums. 

Due to the usage of relatively formal language and low noise, posts from DailyStrength are not subjected to pre-processing. In contrast, tweets contain an approximately equal share of useful information and noise in them. Hence, the tweets are pre-processed by removing URLs, user handles, emoticons, and stopwords. In the case of DailyStrength, because we collect posts from pregnancy-related forums we make the safe assumption that all users posting in the forums are currently pregnant or have been pregnant in the past. However, for Twitter data, manual inspection of a small sample of tweets revealed that approximately 35-40\% of them were false positives (\emph{i.e.}, posts that did not present personal admissions of pregnancy). Therefore, prior to collecting the timelines of the users making the announcement, we employ an automatic text classification technique to further filter out noisy tweets. We manually annotated 1200 randomly selected tweets (approximately 2\% of all the collected tweets) mentioning pregnancy announcements into \emph{isPreg} (legitimate) and \emph{notPreg} (not legitimate) classes.\footnote{This the dataset will be made available with the final version of the paper.} Some examples of pregnancy announcements and their annotations are shown in Table \ref{table:pregclassexamples}. The annotations were performed by two annotators, and the inter-annotator agreement (IAA) for was $\kappa = 0.79$, which is regarded as substantial agreement \cite{landis1977measurement}. Disagreements were resolved by the third author of this paper, who reviewed the disagreement cases and performed the annotations independently. Among the 1200 tweets, 753 tweets were classified as isPreg and 447 were classified as notPreg.

\begin{table}[h!]
  \begin{center}
  \label{table1}
    \caption{Sample tweets showing annotations for pregnancy announcements.}
    \label{table:pregclassexamples}
    \begin{tabularx}{\linewidth}{ X c }
      \toprule
      Tweet & Classification\\ 
      \midrule
         \textit{``I honestly still can't believe I'm almost 5 months pregnant. Like wut."} & isPreg \\
        \textit{``I can't do this anymore. I work my ass off, and I'm eight months pregnant."} & isPreg \\
        \textit{``I'm 18 weeks pregnant today and my 21st birthday is tomorrow. It's a good day"} & isPreg \\ 
           \textit{``I'm 21 weeks, 5 days pregnant. Or, as I like to think of it: 128 days out from reclaiming my spot as my liquor store's favorite customer. "} & isPreg \\ 
        \textit{``It's like just yesterday I was was at the doctor being told I was 14 weeks \& 5 days pregnant .. now I'm 27 weeks \& 3 days"} & isPreg\\
        \textit{``I hate how bloated I get when I'm on my period, like I look like I'm 3 months pregnant"} & notPreg\\
        \textit{``NOW. WAIT. ONE. MINUTE! I'm catching up on \#LHHNY from last night, HOW did Mariah Lynn's mufva go from 5 days pregnant to her 3rd trimester?"} & notPreg\\
        \textit{``I hate that I look at least 4 months pregnant every time I eat something wtf"} & notPreg\\  
        \textit{``Girls will be two days pregnant already posting pictures talking bout ``I'm getting big.""} & notPreg\\  
        \textit{``My sister is five weeks and three days pregnant. I'm going to be an auntie oh my god""} & notPreg\\  
    \end{tabularx}
  \end{center}
\end{table}

Using the annotated data, we perform supervised classification of the tweets. We employ a variant of an existing social media text classification system \cite{sarker2015portable},\footnote{Source code for the classification system is available at: \url{https://bitbucket.org/asarker/adrbinaryclassifier}. Accessed on: 10/10/2016.} which were originally designed for adverse drug reaction detection. Past research on social media text classification suggests that an effective mechanism for classifying short Twitter posts is to generate large numbers of semantic features to balance the sparse word n-gram vectors. Therefore, for the classifier, we primarily remove adverse drug reaction specific features, keep the domain-independent features, and add some additional features. We briefly discuss some of the features in the following paragraphs. 

\subsubsection*{N-grams and synsets}
Word n-grams are the most common text classification features, consisting of sequences of contiguous \emph{n} words in a text segment. We preprocess the texts by performing stemming and lowercasing, and use 1-, 2-, and 3-grams as features.

In addition to the words themselves, we use their synonyms in some cases to increase vocabulary coverage. For each adjective, noun or verb in a tweet, we use WordNet. \footnote{\url{http://wordnet.princeton.edu/}. Accessed on: 01/05/2016.} to identify the synonyms of that term and add the synonymous terms as features.



\subsubsection*{Sentiment representing features}
Our inspections of the announcements suggest that users generally express strong sentiments when making pregnancy announcements. So, we add features that express the sentiments of the users in various scales. We assign three sets of scores to sentences based on three different measures of sentiment. The first set of scores are derived from lists of positive and negative terms \cite{hu2004mining}, the second set of scores are dependent on the prior polarities of terms present in a post \cite{guerini2013sentiment}, and the third set of scores are derived from a subjectivity lexicon that presents both polarity and subjectivity \cite{wiebe2005annotating}. 

\subsubsection*{Word clusters}
Recent research on social media based text classification suggests that using generalized representations of words, such as clusters of similar words, may improve performance \cite{nikfarjam2015pharmacovigilance}. In our work, we use the clusters generated by Owoputi \emph{et al.} \cite{owoputi2012part}. The authors generate the clusters by first learning vector representations of words \cite{mikolov2013efficient} from over 56 million tweets, and then employing a Hidden Markov Model-based algorithm that partitions words into a base set of 1000 clusters, and induces a hierarchy among those 1000 clusters.\footnote{The clusters are publicly available at: \url{http://www.cs.cmu.edu/~ark/TweetNLP/}. Accessed on: 1/24/2017.}

To generate features from these clusters, for each tweet, we identify the cluster number of each token, and use all the cluster numbers associated with a tweet in a bag-of-words manner. Thus, every tweet is represented by a set of cluster numbers, with semantically similar tokens having the same cluster number. 

\subsubsection*{Classification}
Using these features, we trained Support Vector Machine (SVM) classifiers for the classification task. We used an RBF kernel, and we optimized the value of the cost parameter via 10-fold cross-validation over the 1200 annotated posts.\footnote{We used the LibSVM implementation packaged with the python scikit-learn implementation: \url{http://scikit-learn.org/}. Accessed on: 10/11/2016.} We obtained the best results with cost=64.0, and we used this setting to classify all the identified tweets in our collection. Results of the classification are presented in the next section, including classification performance and the number posts (Table \ref{table:tweetlegpregclassif}).

\subsection{User Health Timeline Extraction}
After the classification step, all the handles of the users classified to be legitimately pregnant are identified, and we attempt to collect their other posts using the Twitter API. We index all the posts into Apache Lucene\footnote{\url{https://lucene.apache.org/}. Accessed on: 11/23/2015.} for further analysis. Using the API, we collect all the user posts that are available from the past (\emph{i.e.}, up to the limit allowed by Twitter) and sort them in chronological order, and we continue collecting tweets over time to monitor future health-related events. 


For DailyStrength, however, since the forums we chose are all pregnancy-related, it is assumed that all users posting in these  forums are/have been pregnant at some point during their membership to the website. The users can post across different forums on the website which can include interesting information such as drug intake admissions and adverse events. Hence, for each user posting a comment in one of the pregnancy related forums, we collect the user's posts in all available forums to construct their timelines. Finally, we index each individual timeline into Apache Lucene with the following fields: userid, time, text, and trimester (if available) for further processing.



\subsection{Timeline Analysis}

Using the collected timelines, we attempt to explore if and how health-related events can be clustered into coarse-grained temporal windows. The duration of a pregnancy may be divided into three trimesters: first-- week 1 through week 12, second-- week 13 through week 27, and third-- week 28 through birth. To successfully identify the trimester associated with a posted health-related event, information about the pregnancy start date is required. Via our manual inspections of the timelines, we discovered that pregnant mothers who announce their pregnancies over Twitter also often provide clues about the progress of the pregnancies. Consider the tweets below:

\begin{quote}
\textit{Oh well managed 8 out of 10 combat tracks, not bad at 28 weeks pregnant with the flu but still disappointing \#frustrated}

\textit{I'm officially 20 weeks pregnant \& I've also never felt more sick in my life.:)}
\end{quote}
The first tweet was posted during the third trimester and the second tweet was posted during the second trimester of pregnancy. Using this information, and the timestamp of the tweets, all the posts within a timeline can be grouped into the three trimesters. The key NLP challenge in this problem is to detect the statements regarding the progress of the pregnancies.

We use a combination of term and pattern matching algorithms to detect these trimester identifiers in each timeline. However, for Twitter, due to the 3200 tweet limitation enforced by the API, not all timelines that are extracted have all the tweets posted during the pregnancy time period. In our current algorithm, we first attempt to identify all tweets that mention the terms `\emph{pregnant}' and `\emph{pregnancy}' (seed word). Next, terms within a specified context window of the seed word are collected. Based on the empirical assessment, we settled for symmetric context window of size 6 terms. Within the context window, the algorithm then searches for \emph{key} temporal terms such as `\emph{week}' and `\emph{month}', along with the presence of a number mention (\emph{e.g.}, six, 12, eighteen and so on). The number, along with the other mentioned terms are extracted and compared to the timestamp of the associated tweet to identify the trimester.

Following the organization of the timelines into trimesters, we assessed, in a preliminary fashion, if trimester-specific health events can be collected for further analysis. Depending on the intent of a study, the type of information that requires mining may vary, and detailed trimester-based health-related event analysis is outside the scope of this paper. Therefore, we simply focused on generating frequencies of the drugs that are mentioned at each trimester to make rough estimates about the drug usage patterns of the cohort at each phase. 
We perform a keyword search for each drug in Apache Lucene to obtain the drug mentions by users. Here, we make an assumption that all drug mentions are admissions of drug intake by the user. We query our Lucene index, and, for each drug, compute the number of users who have consumed it. 
The goal was to ascertain if a drug-usage information is available, rather than to perform a thorough analysis, which we leave as future work. Distributions of the drug mentions are presented in next section.


\section{Results and Discussions}
\begin{figure}
 \centering
 \includegraphics[width = 0.52\textwidth]{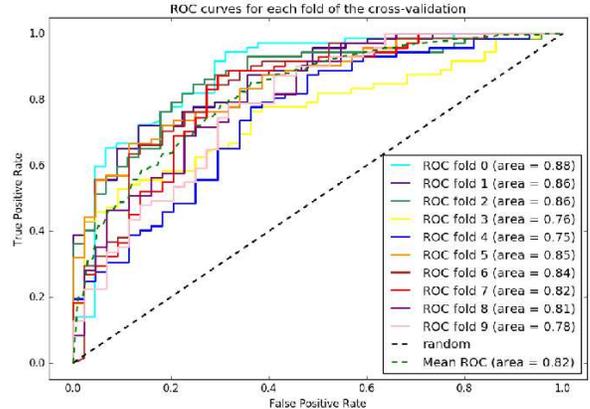} 
 \caption{ROC curves for each fold of the 10-fold cross-validation, and the mean. }
 \label{roc_curve}
 \end{figure}

The performance of our classifier was evaluated via 10-fold cross-validation, and the best results obtained are presented in Table \ref{table:tweetlegpregclassif}. We compared the performance of the SVM to that of a Na\"ive Bayes baseline, which obtained an F-measure of 0.70 for the isPreg class. Figure \ref{roc_curve} shows the ROC curves for each of the 10-folds of cross-validation, including the mean ROC for the positive class. The area under the mean ROC curve is 0.82. Running the SVM classifiers on our collected data resulted in the discovery of 34,895 legitimate pregnant women from a total of 53,820 users.  
 
\begin{table}[h!]
 \begin{center}
   \caption{Results from tweet classification for legitimate pregnancy announcements using SVM}
   \label{table:tweetlegpregclassif}
   \begin{tabular}{lccc}
     \toprule
     Classification Result & Precision & Recall & F-measure\\ 
     \midrule
        isPreg & 0.83 & 0.79 & 0.81 \\
        notPreg & 0.84 & 0.77& 0.80 \\ 
        \bottomrule
   \end{tabular}
 \end{center}
\end{table}

We applied our pregnancy trimester extraction algorithm on the 34,895 user timelines classified as legitimate pregnant users. Our algorithm detected pregnancy time-period for 15,523 (approximately 45\%) users and was able to further categorize each tweet belonging to these timelines into one of the three trimesters. The remaining user handles were discarded from the analysis performed in the rest of this paper. We were able to collect over 30 Million tweets from these 15,523 users. Table \ref{table:healthtimelineexample2} showcases a user timeline with the pregnancy trimester details and health-related tweets in each of the three trimesters. 

\begin{table*}[h!]
  \begin{center}
    \caption{Excerpts from a Twitter user's health timeline in reverse chronological order}
    \label{table:healthtimelineexample2}
    \begin{tabularx}{\linewidth}{l X c c }
      \toprule
      No & Tweet & Trimester & Category\\ 
      \midrule
\rownumber & \textit{``God bless Zantac for helping me not want to throw up from the insane heartburn I've been having. \#pregnancyproblems \#37weeks"} &    third &    drug, condition \\
\rownumber & \textit{``Ugh. Awful dreams; was obviously clenching my teeth all night - woke up with sore jaw; a headache. \#ouch \#tired \#iwanttogobacktobed"} &    third &    condition \\
\rownumber & \textit{``Went to the chiropractor today and my lower back actually feels worse right now. \#sore"} &    third &    condition \\
\rownumber & \textit{``If I don't drink Powerade before bed, I get awful leg cramps. If I do, I get awful heartburn. \#cantwin \#PregnancyProblems"} &    third &    condition \\
\rownumber & \textit{``Betting I'll be back at Dr. AnonX's again real soon. Anonymized has a 101.8 degree fever. Why do my kids get sick so much?! \#mommyproblems \#sickkids"} &    third &    condition \\
\rownumber & \textit{``W/ the leg pain from the INSANE cramp I had this AM; the pelvic pain from how baby's laying, the waddle is strong today. \#pregnancyproblems"} &    third &    condition \\
\rownumber & \textit{``Fell asleep around 10:30. Just woke up sweating and uncomfortable. It's too hot in here to sleep now. \#PregnancyProblems"} &    third &    condition \\
\rownumber & \textit{``Lunch was absolutely AMAZING (Tuscan chicken; artichoke soup from @anonymized) but now I have the worst heartburn. \#PregnancyProblems"} &    third &    condition \\
\rownumber & \textit{``Baby's moving around like crazy; the pain in my side/back is FINALLY letting up. RELIEF! \#pregnancyproblems"} &    third &    condition \\
\rownumber & \textit{``My pelvis hurts so freaking bad right now and I still have 14 weeks to go til my due date. \#PregnancyProblems"} &    second &    condition \\
\rownumber & \textit{``At the doctor with Anon again. Fever was 102 this AM, gave her Tylenol, then up to 104 after her nap. Her only complaint - being cold."} &    second &    drug, condition \\
\rownumber & \textit{``@anonymized Taking multivitamin gummy in AM, calcium + D in afternoon, prenatal at bedtime....just like I always have."} &    second &    drug \\
\rownumber & \textit{``Well this is a new one - my Vitamin D level is actually too HIGH. Now OB wants me to see bariatric doc/nutritionist again."} &    second &    drug \\
\rownumber & \textit{``It'll be another sleepless night checking on Anon periodically. 104.8 degree fever earlier, Motrin brought it down to 100.1"} &    second &    drug \\
\rownumber & \textit{``So help me if I'm getting ANOTHER cold I'm gonna be pissed. Scratchy sore throat and runny nose all of a sudden."} &    second &    condition \\
\rownumber & \textit{``Slim chance it's the start of appendicitis. If I get a fever, nausea/vomiting/diarrhea,; pain is worse, I need to come back ASAP."} &    second &    condition \\
\rownumber & \textit{``Weak gag reflex + coughing + snot = disaster waiting to happen. I just want to go home, crawl in bed, and sleep until this cold is gone."} &    second &    condition \\
\rownumber & \textit{``Not bring able to take anything for this stupid cold is awful. So miserable. \#stuffedup \#cantbreathe \#cough"} &    second &    condition \\
\rownumber & \textit{``It's amazing how a headache, raging hormones; lack of sleep make you want to stuff a pillow in the face of a snoring husband. \#shutup"} &    second &    condition \\
\rownumber & \textit{``For the second time in a row, Tylenol PM has left me wide awake at 3AM after passing out for a whopping 5 hours. \#needmoresleep"} &    second &    drug \\
\rownumber & \textit{``I love the Olympics! Too bad I just took some Tylenol PM that's starting to kick in so I won't be watching much longer tonight."} &    second &    drug \\
\rownumber & \textit{``AnonT stayed home from work today on daddy duty so I just popped some Tylenol PM and I'm sleeping this cold away. \#goodnight"} &    first &    drug \\
\rownumber & \textit{``My throat is dry; sore from breathing through my mouth but breathing through my nose isn't possible right now. \#snot \#sick"} &    first &    condition \\
\rownumber & \textit{``Woke up to anonymized crying around 5:30; a skull-knocker headache. And guess what...no Tylenol except for PM stuff. \#crap"} &    first &    drug \\
\rownumber & \textit{``Been sleeping awful so AnonT said to take Tylenol PM; get a good nights sleep. Come up to bed; discover we're out of Tylenol PM. \#gofigure"} &    first &    drug, condition \\
\rownumber & \textit{``Finally start to feel drowsy so I try to sleep, but then I get all twitchy and can't lay still. \#insomnia \#sleepproblems"} &    first &    condition \\
\rownumber & \textit{``Pseudo gallbladder attacks in the middle of the night are awesome, said no one ever. \#pain"} &    first &    condition \\
\rownumber & \textit{``Just turned my head and something popped in my neck the wrong way. Big ouch. Need to hit up the chiropractor on my way home. \#hurt"} &    first &    condition \\
\rownumber & \textit{``Gotta pack up the kids; head to MQT. Woke up with a skull knocker headache so me thinks it's time for a back; neck cracking. \#chiropractor"} &    first &    condition \\
\bottomrule    
  \end{tabularx}
  \end{center}
\end{table*}

\begin{figure*}
 \centering
 \includegraphics[width = 0.90\textwidth]{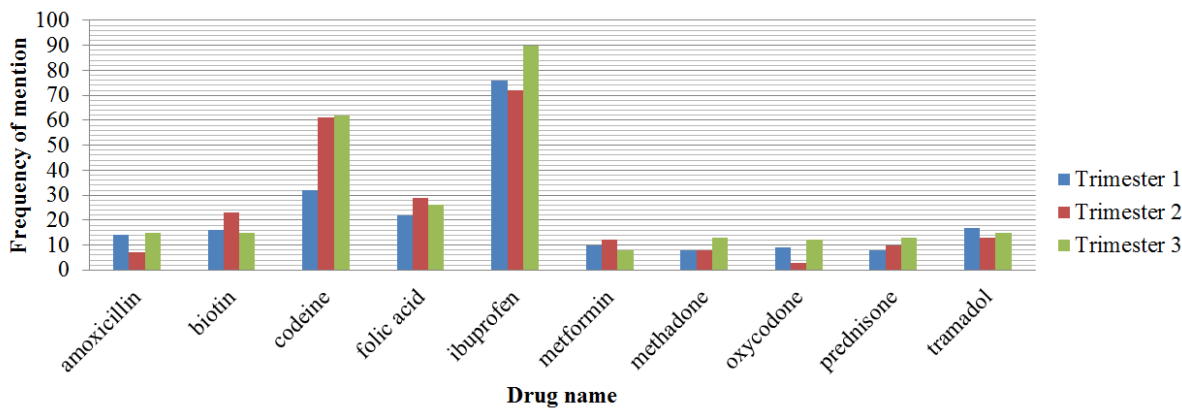} 
 \caption{Distributions of top 10 drug mentions within the timelines across the three trimesters.}
 \label{fig:drugbytrimesters}
 \end{figure*}

We observe that the timeline contains health-related information such as drug intakes (in rows 1, 12, 20, 21, 22) and conditions/events (\emph{e.g.}, rows 1, 2, 4, etc.). We also notice that a large proportion of drug and condition mentions happen to be first-hand experiences. However, not all mentions of drugs are intakes (rows 24 and 25) and not all drug intakes are drug intakes by the user (rows 11 and 14). Similarly, not all conditions mentioned in the tweets are experienced by the user (rows 5 and 11). Mining drug intake and events are important in pharmacovigilance research for tracking ADRs. Hence, accurately distinguishing personal drug intake and events from mentions is an important NLP challenge that we intend to address in the future.

Trimester detection adds a very interesting NLP challenge. While some tweets are relatively easy to detect and were successfully processed by our rule-based algorithm, we found some that were missed or mis-classified. Consider the following Tweets, for example:
\begin{quotation}
\textit{I b getting so much pressure next week is gone b my last week pregnant who want to make a bet lol}

\textit{It is crazy to me that I am only 3 days past 13 weeks pregnant.} 
\end{quotation}
Our approach currently fails to detect the first tweet and mis-classifies the second tweet as first trimester instead of second. We leave the optimization of our detection algorithm as future work.

Figure \ref{fig:drugbytrimesters} shows the distribution of popular drug mentions across the pregnancy trimesters for Twitter users. Even the most common drugs were mentioned by less than 0.5\% of the users and the proportion of actual intakes may be lower. For instance, ibuprofen was one of the most common drugs mentioned in user timelines and it was mentioned by 76 unique users in their first trimester, 72 in their second, and 90 in their third. For a collection of more 15,000 user timelines, we find this proportion of mentions to be low for extensive analysis and hence we intend to expand our search terms and algorithms for cohort selection in the future.


For DailyStrength, our timeline collection approach retrieved a total of 257,531 posts from 11,435. In contrast to tweets, which are restricted to 140 characters, DailyStrength posts are longer. Thus, tracking the progress of pregnancies from their announcements to derive trimester information requires further NLP research. Discovering drug intake, however, is similar, and we find that common drug mentions within the user timelines in DailyStrength include drugs such as folic acid, aspirin, zoloft and tylenol.

\section{Limitations and Future Work}
We intend to build on this preliminary work in several key areas. 
Employing more sophisticated outcome detection techniques is an important future goal of this study. From the NLP perspective, our technique does not take into account tense (\emph{e.g.}, past/present) and so the chronological order of posts may not represent the chronological ordering of events. Also, no mechanism is applied for gender detection among pregnancy announcement tweets, although our self-admission classifier does attempt to ensure that users included in the cohort are genuinely pregnant themselves. 

Among other things, we intend to expand the drug list by including misspellings, spelling variations, phonetic variations and abbreviations of each drug. Similar to drug usage pattern extraction, we could use a disease and disorder extraction method to classify mentions of diseases which would explain the reason why certain individuals consume a particular drug. As mentioned already in the paper, our trimester detection technique is currently not optimal, and we will improve it via the addition of more rules. Since the performance and effectiveness of the downstream applications and analyses depend heavily on the data collection and classification steps, our immediate focus will be to improve these. We will employ more queries to significantly increase the size of the cohort, and improve the performance of the classification step via the annotation of a much larger data set and the application of more sophisticated classification techniques. Ensembles of classifiers have been shown to perform particularly well for complex text classification tasks \cite{sarker15ebm}, and we will attempt to develop such systems with the view of maximizing recall while maintaining high precision. 



\section{Conclusion}
In this paper, we presented the novel idea of collecting longitudinal health-related information about targeted cohorts from social media. We focused on the cohort of pregnant women in this study--- a group that is not included in pre-market clinical trials. We presented a pipeline which includes three stages--- identification of cohort, collection, and analysis. We discovered that large numbers of pregnant women can be identified with high-precision via a combination of rule-based and machine learning techniques. We discussed how health-related timelines can be gathered from two different social networks. Finally, we showed how temporal categorization of the timeline may be performed, and we verified that trimester-specific health-related information can be mined from the pre-processed timelines. 

We discussed several limitations of our work, which will be addressed in future research. Crucially, while we focused solely on one cohort, our pipeline can be generalized for other population groups as well. This form of analysis may be particularly useful for population groups about whom data may not be available from other sources. In addition, social media may reveal information that people may not generally share via other means (\emph{e.g.}, drug abuse/usage of illicit drugs). The results obtained by our current work are very promising and warrant future research.

\bibliographystyle{abbrv}
\bibliography{sigproc}  
\end{document}